\documentclass[10pt,twocolumn,letterpaper]{article}

\usepackage{cvpr}
\usepackage{times}
\usepackage{epsfig,epstopdf}
\usepackage{graphicx}
\usepackage{amsmath}
\usepackage{amssymb}
\usepackage{color}
\usepackage{colortbl}
\usepackage{subfigure}
\DeclareMathOperator*{\argmin}{\arg\!\min}
\usepackage{flushend}

\cvprfinalcopy % *** Uncomment this line for the final submission

\def\cvprPaperID{1839} % *** 

\ifcvprfinal\fi
\begin{document}

%%%%%%%%% TITLE
%\title{Forecasting Campus Dynamics: a Multi-class Approach}

%\title{Learning Social Navigation in a Crowded Multi-Class Setting}

\title{Forecasting Social Navigation in Crowded Complex Scenes}

\author{Alexandre Robicquet, Alexandre Alahi, Amir Sadeghian,\\ Bryan Anenberg, John Doherty, Eli Wu, and Silvio Savarese\\
Stanford University\\
{\tt\small arobicqu@stanford.edu}
}

\maketitle
%\thispagestyle{empty}

% \thispagestyle{empty}

%%%%%%%%% ABSTRACT

\begin{abstract}

When humans navigate a crowed space such as a university campus or the sidewalks of a busy street, they follow common sense rules based on social etiquette.   In this paper, we argue that in order to enable the design of new algorithms that can take fully advantage of these rules to better solve tasks such as target tracking or trajectory forecasting, we need to have access to better data in the first place. To that end, we contribute the very first large scale dataset (to the best of our knowledge) that collects images and videos of various types of targets (not just pedestrians, but also bikers, skateboarders, cars, buses, golf carts) that navigate in a real world outdoor environment such as a university campus.   We present an extensive evaluation where different methods for trajectory forecasting are evaluated and compared. Moreover, we present a new algorithm for trajectory prediction that exploits the complexity of our new dataset and allows to: i) incorporate  inter-class interactions into trajectory prediction models (e.g, pedestrian vs bike) as opposed to just intra-class interactions (\textit{e.g.}, pedestrian vs pedestrian); ii) model the 
degree to which the social forces are regulating an interaction. We call the latter "\textit{social sensitivity}" and it captures the “sensitivity” to which a target is responding to a certain interaction. An extensive experimental evaluation demonstrates the effectiveness of our novel approach.

\end{abstract}

\section{Introduction}

When pedestrians walk in a crowded space such as a university campus, a shopping mall or the sidewalks of a busy street, they follow common sense conventions based on social etiquette. For instance, they would yield the right-of-way at an intersection as a bike approaches very quickly from the side, avoid walking on flowers, and respect personal distance. By constantly observing the environment and by navigating through it, humans have learnt the way other humans typically interact with the physical space as well as with the targets that populate such spaces —\textit{e.g.}, humans, bikes, skaters, electric carts, cars, toddlers, etc. They use these learned principles to operate in very complex scenes with extraordinary proficiency.

Researchers have demonstrated that it is indeed possible to model the interaction between humans and their surrounding environment to improve or solve numerous computer vision tasks: 
for instance, to make pedestrian tracking more robust and accurate \cite{yamaguchi2011you,pellegrini2010improving,leal2014learning,choi2012unified,smeulders2014visual},
to enable the understanding of activities performed by groups of individuals 
\cite{xie2013inferring,choi2014understanding,lan2010beyond,choi2009they}, to enable  accurate  prediction of target trajectories in future instants \cite{kitani2012activity,lerner2007crowds,trautman2013robot,cucchiara2004probabilistic}. Most of the time, however, these approaches operate under restrictive assumptions whereby the type and number of interactions are limited or the testing environment is often contrived or artificial.

\begin{figure}[t]
\begin{center}
   \includegraphics[width=0.9\linewidth]{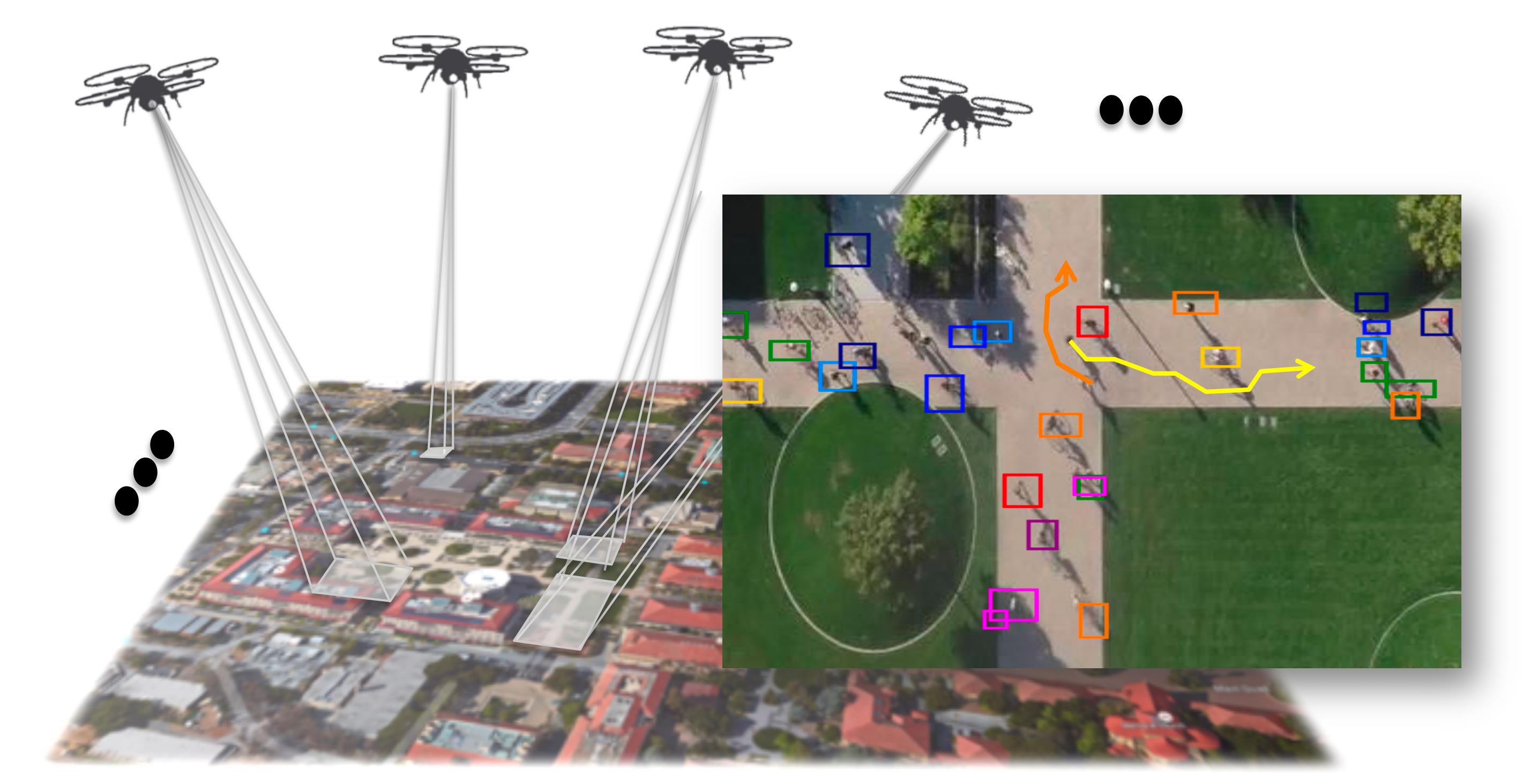}
\end{center}
   \caption{We aim to forecast human social navigation in a multi-class setting where pedestrians, bicyclists, skateboarders and carts share the same space. We hence have collected a new dataset with a quadcopter flying over more than 100 different crowded campus scenes.}
\label{fig:pull}
\end{figure}

In this paper, we argue that in order to learn and use models that allow mimicking, for instance, the remarkable human capability to navigate in complex and crowed scenes, we need to have access to better data in the first place. To that end, we contribute the very first large scale dataset (to the best of our knowledge) that collects images and videos of various types of targets (not just pedestrians, but also bikes, skateboarders, cars, buses, golf carts)
%[[LIST all of them]]) 
that navigate in a real world outdoor environment such as a university campus. Our dataset comprises of more than 100 different top-view scenes for a total of 20,000 targets engaged in various types of interactions. Target trajectories along with their target IDs are annotated which makes this an ideal testbed for learning and evaluating models for multi-target tracking, activity understanding and trajectory prediction at scale.

Among all the problems discussed above, in this paper we are interested in focusing on the latter — target trajectory forecasting from videos — whereby the ability to comply to social etiquettes and common sense behavior is critical.  In particular, we believe that our new dataset creates the opportunity to generalize state-of-the-art methods for trajectory forecasting and evaluate them on a more effective playground. While two leading families of methods for target trajectory forecasting (social forces \cite{hughes2003flow,helbing1995social,yamaguchi2011you,pellegrini2010improving} and gaussian process \cite{boyle2005dependent,tay2008modelling,trautman2013robot} ) have shown promising results on existing datasets \cite{pellegrini2009you,lerner2007crowds},
%[ add ref to ETH, HOTEL, etc…], 
they have never been tested at scale and in real-world scenarios where multiple classes of targets are present (i.e., not just pedestrian but also cars, bikes, etc.) as part of a complex ecosystem of interacting targets. In this work, we are providing an answer to these questions as well as contributing a generalization of these methods that allows: 

(i) Incorporation of inter-class interactions into trajectory prediction models (e.g, pedestrian vs bike) as opposed to just intra-class interactions (\textit{e.g.}, pedestrian vs pedestrian). For instance, a pedestrian would speed up or slow down his/her pace if a bike is rapidly approaching from the side, whereas his/her pace wouldn't change much of a pedestrian (instead of a bike) is rapidly approaching from the same direction.

(ii) Modelling the degree to which the social forces are regulating an interaction. We call this \textit{social sensitivity} and it captures the “sensitivity” to which a target is responding to a certain interaction. A low \textit{social sensitivity}  means that a target motion is not affected much by other targets that are potentially interacting with it. A high \textit{social sensitivity}  means that the target navigation is highly dependent on the position of other targets. We model \textit{social sensitivity} in a data driven fashion and introduce it as a latent variable in the forecasting model. The introduction of the \textit{social sensitivity} variable increases the flexibility in characterizing various modalities of interactions - for instance, some pedestrians may look more “aggressive” while walking because they are in rush whereas others might show a milder behavior because they are just enjoying their walk.

We present an extensive experimental evaluation analysis that compares various state-of-the-art baselines on the newly proposed dataset, and demonstrates that our generalized models based on on \textit{social sensitivity} and multiple classes of target behaviors enable better prediction performance than baseline methods that assume that all the targets belong to the same class.

\section{Previous Work}

A large variety of methods has been proposed in the literature to describe, model and predict human behaviors in a crowded space. For instance, Antonini \textit{et. al.} use the Discrete Choice Model to synthesize human trajectories in crowded scenes \cite{antonini2004discrete,antonini2006discrete}. Other methods learn motion patterns by clustering trajectories \cite{hu2007semantic,lerner2007crowds, morris2011trajectory,kim2011gaussian}. In this work, we focus on two distinct methods of resolution: energy minimization based, and probabilistic ones.

\paragraph{Energy Minimization.} An exhaustive study of crowd analysis is introduced in \cite{treuille2006continuum} by Treuille \textit{et. al.}. They focus on  real-time synthesis of crowd motion for thousands of individuals with intersecting paths.  The human motion is viewed as a per particle energy minimization, and adopts a continuum perspective on the system. This formulation yields a set of dynamic potential and velocity fields over the domain that guides all individual motions simultaneously, and is designed for large groups with common goals, not for scenarios where each person's intention is distinctly different. 

%Virtually all previous work has been target-based, meaning that motion is computed separately for each individual. But it is difficult to develop behavioural rules that consistently produce realistic motion. Global path planning for each target quickly becomes computationally expensive, particularly in real-time contexts. Moreover, local path planning often results in myopic, less
%realistic crowd behavior. 
%Our approach unifies global path planning and local collision avoidance into a single optimization framework.

The most popular method is the \textit{Social forces} model first introduced by \textit{D. Hellbing} and \textit{P. Molnar} in \cite{helbing1995social}. It has been extensively studied in robotics \cite{luber2010people} and for tracking algorithms in computer vision \cite{yamaguchi2011you,pellegrini2010improving,mehran2009abnormal,leal2011everybody,alahi14,kretzschmar2014learning,yi2015understanding}. Pedestrians react to energy potentials caused by the interactions with other targets  and static obstacles through forces (repulsion or attraction). Our proposed method is an extension of the Social Forces model (more details are provided in Section \ref{momodel}).  %This paper presents a new type of crowd simulator driven by dynamic potential fields which integrate both global
%navigation and local collision avoidance into one framework.\\

\paragraph{Probabilistic Forecasting} A large body of work is based on \textit{Inverse Reinforcement Learning} \cite{kitani2012activity,ziebart2009planning,henry2010learning}. The key idea underlying this family of techniques is to learn a reward (or cost) function that best explains the final decisions \cite{ziebart2008maximum}. Reward functions are represented by log-linear functions of features describing a task environment. While these techniques have been shown to work extremely well in several applications \cite{levine2011nonlinear,ziebart2008maximum,thompson2009probabilistic}, they assume that all feature values are known and static during each demonstrated planning cycle. %In contrast, our scenario requires learning from example,paths that are the result of a (simulated) person updating feature value estimates and re-planning on the fly. 
%On te one hand, \cite{henry2010learning} introduce a new method for Inverse Reinforcement Learning based on the maximum entropy and the use of local estimation through Gaussian Processes. On the other hand, \cite{kitani2012activity} present a more intuitive way to tackle our problem, by modeling the interaction between moving targets and semantic perception of the environment. We also can discover in \cite{ziebart2009planning}  a novel approach for predicting the future trajectory of a person moving through an environment using a soft-max version of goal-based planning.\\
These approaches mostly model motion trajectories as transitions between discretized states. The main disadvantage of discretization is the need to determine the discretization of the state spaces and the association of observations to these state spaces.

Wang \textit{et. al.}
 \cite{wang2008gaussian} and Tay \textit{et. al.} \cite{tay2008modelling} introduced new approaches to predict paths using Gaussian processes. The typical motion paths are smooth and avoid the problems associated with discretization and the representation of motion paths with Gaussian process lend itself naturally to clustering using the Gaussian mixture model.  Gaussian processes makes it possible to represent paths
as continuous functions in a probabilistic manner. The problem of discretization is conveniently side stepped and prediction on the future path taken can be performed in a theoretically proper probabilistic framework \cite{trautman2013robot}. 

In general, most of existing approaches operate under restrictive assumptions whereby the testing environment is often contrived to a single class of dynamic. In the next sections, we present our initiative to cope with such limitation.

\section{Campus Dataset}

We aim to learn the remarkable human capability to navigate in complex and crowded scenes. 
%Modelling and learning human navigation from the data depends on the informativeness of the used datasets. 
%In practice, urban scenes are made of multiple classes of targets interacting with each other. 
Existing datasets mainly capture the behavior of humans in spaces occupied by a single class of object, \textit{e.g.}, pedestrian-only scenes \cite{pellegrini2009you,lerner2007crowds,alahi14}. However, in practice, pedestrians share the spaces with other classes of objects such as bicyclists, or skateboarders to name a few. For instance, on university campuses, a large variety of these objects interacts at peak hours. We want to study social navigation in these complex and crowded scenes occupied by several classes of objects.  

To the best of our knowledge, we have collected the first large-scale dataset that has images and videos of various types of targets interacting in a real-world university campus. Our dataset captures the following types of interactions:
\begin{itemize}
\item target-target interactions, \textit{e.g.}, a bicyclist avoiding a pedestrian, 
\item target-space interactions, \textit{e.g.}, a skateboarder turning around a roundabout.
\end{itemize}

\begin{table}[!ht]
\begin{center}
\footnotesize{
\begin{tabular}{|c|c|c|c|c|}
\hline Dataset & Frames & Targets  & Interactions & Physical class\\ 
\hline \textsc{Isengard} & 134079 & 2044 &  6472& 6\\ 
%\hline \textsc{Little} & 46764 & 1130 &  ?&5\\
\hline \textsc{Hobbiton} & 138513 & 3821 &  14084 &6\\
\hline \textsc{Edoras} & 47864 & 1186 &  4684&5\\
\hline \textsc{Mordor} & 139364 & 4542 &  68459& 6 \\ 
%\hline \textsc{Main Quad} & 2036 & 64 & 152 & 2 \\ 
\hline \textsc{Fangorn} & 249967 & 3126 &  45520&6 \\ 
\hline \textsc{The Valley} & 219712 & 4845 &46062  & 6 \\ 
\hline \textsc{Total} & 929499 & 19564 & 185281 & 6 \\ 
% \hline \textsc{Bytes} & 18050 & 46 &  & 4\\ 
%\hline \textsc{Campus} & \textbf{\textcolor{black}{20458}} & \textbf{\textcolor{black}{752}} & \textbf{\textcolor{black}{28250}}&\textcolor{black}{\textbf{7}}\\ 
\hline 
\end{tabular}
}
\end{center}
\caption{Our campus dataset characteristics. We group the scenes and refer to them using fictional places from the "Lord of the Rings".} \label{mapcampus}
\end{table}

\paragraph{Target-target interactions} We say that two targets interact when their collision energy (described by Equation \ref{eq5})  is non-zero, \textit{e.g.}, a pedestrian avoiding a skateboarder. These interactions involve multiple physical classes of targets (pedestrians, bicyclists, or skateboarders to name a few), resulting into 185K annotated target-target interactions. We intentionally collected data at peak hours (between class breaks in our case) to observe high density crowds. For instance, during a period of 20 seconds, we observe in average from 20 to 60 targets in  a scene (of approximately $900m^2$).
 
\begin{figure*}[t]
\begin{center}
   \includegraphics[width=0.9\linewidth]{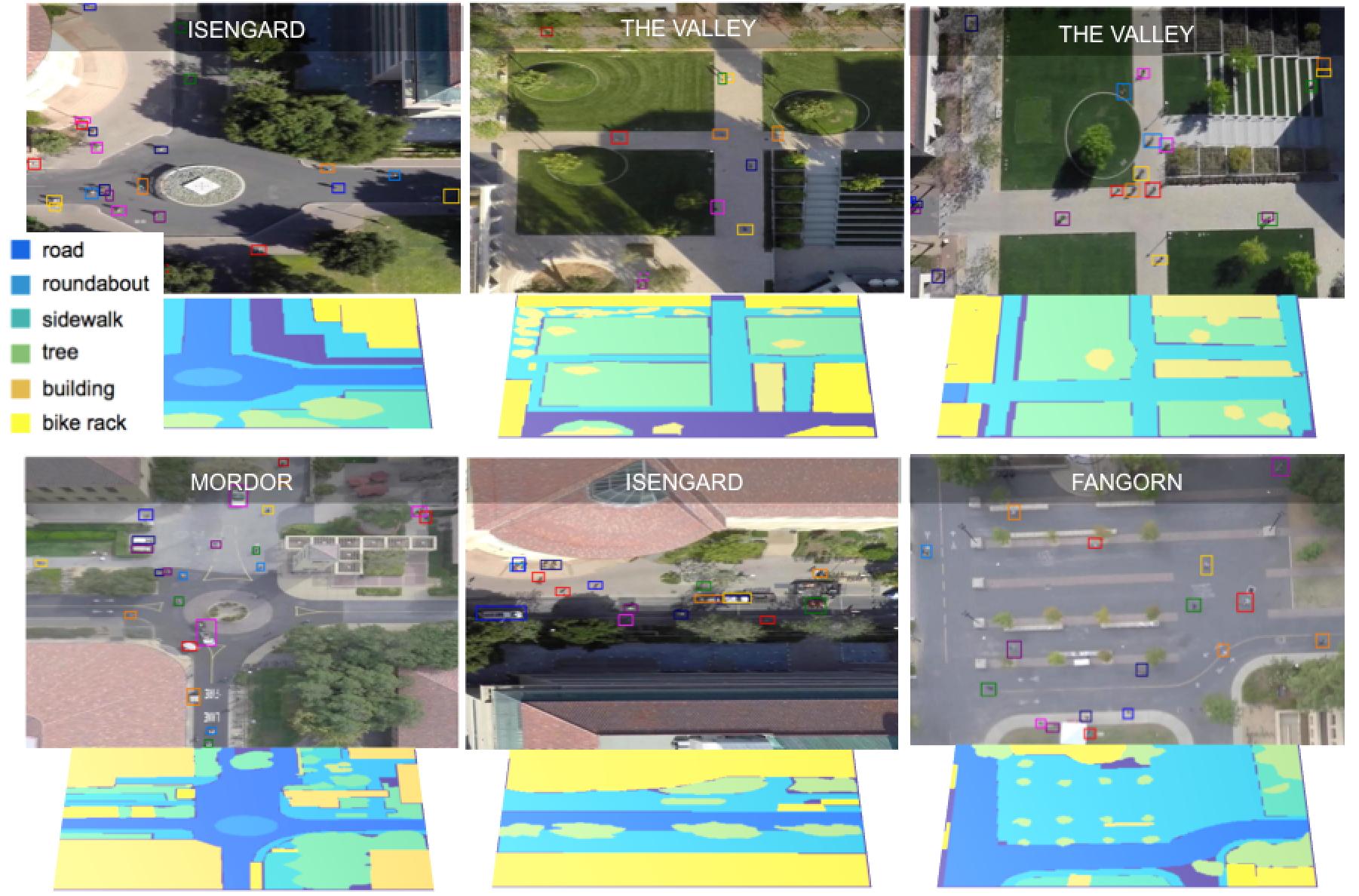}
\end{center}
   \caption{Some examples of the scenes captured in our dataset. We have annotated all the targets (with bounding boxes) as well as the static scene semantics (rows 2, 4, and 6). The color codes associated to target bounding boxes represents different track IDs. }
\label{fig:sfig1}
\end{figure*}

\paragraph{Target-space interactions.} We say that a target interacts with the space when its trajectory deviates from a linear one in the absence of other targets in its surrounding, \textit{e.g.}, a skateboarder turning around a roundabout. To further analyze these interactions, we also labeled the scene semantics of more than 100 static scenes with the following labels: road, roundabout, sidewalk, grass, building, and bike rack (see Figure \ref{fig:sfig1}).  We have approximately 40k ``target-space" interactions. 

To the best of our knowledge, it is the first dataset to depict complex interactions at such a scale. Tables \ref{mapcampus} and \ref{mapcampus2} present more details on our collected dataset. The scenes are grouped into 6 areas based on their physical proximity on campus. The dataset comprises more than 19K targets consisting of 11.2K pedestrians, 6.4K bicyclists, 1.3k cars, 0.3K skateboarders, 0.2K golf carts, and 0.1K buses. 

Each scene is captured with a 4k camera mounted on a quadrotor platform hovering above various intersections on a University campus at an altitude of approximately eighty meters. The videos are also available for further research in detection, recognition, tracking from UAV data. The videos have been processed (\textit{i.e.} undistorted and stabilized), and annotated with their class label and their trajectory in time and space is identified. 

Our dataset can be used to conduct research in activity and scene understanding. For example, the collected trajectories can be used to infer the functionality map of a scene \cite{Gupta2009,Xie2013,Zhu2014,Saxena2015}, \textit{e.g.}, infer sitting areas, and improve image segmentation.
We envision our dataset to be an ideal testbed for pushing the limits of visually intelligent machines. It enables the design of new methods that allow learning multi-target interactions at a large scale as well as pushing research on multi-target tracking. 

\begin{table}[!ht]
\begin{center}
\footnotesize{
\begin{tabular}{|c|c|c|c|c|c|c|c|}
\hline Dataset & Bi & Ped  & Skate & Carts & Car & Bus\\ 
\hline \textsc{Isengard} & 1004 & 926 &  57 & 19 & 23 & 15\\ 
%\hline \textsc{Little} & 639 & 475 &  5 & 0 & 4 & 7\\
\hline \textsc{Hobbiton} & 163 & 2493 &  24 & 18 & 1065 & 58\\
\hline \textsc{Edoras} & 224 & 956 &  2& 2 & 2 & 0\\
\hline \textsc{Mordor} & 2594 & 1492 &  111 & 154 & 165 & 26\\ 
%\hline \textsc{Main Quad} & 8 & 56 &  0 & 0  & 0 & 0\\ 
\hline \textsc{Fangorn } & 1017 & 1991 &  50 & 30  & 27 & 11\\ 
\hline \textsc{The Valley} & 1362 & 3358 & 89 & 21 & 10 & 5 \\ 
\hline \textsc{Total} & 6364 & 11216 & 333 & 244 & 1292 & 115 \\ 

% \hline \textsc{Bytes} & 8 & 35 & 2 & 0 & 1 & 0\\ 
%\hline \textsc{Campus} & \textbf{\textcolor{black}{20458}} & \textbf{\textcolor{black}{752}} & \textbf{\textcolor{black}{28250}}&\textcolor{black}{\textbf{7}}\\ 
\hline 
\end{tabular}
}
\end{center}
\caption{Details on the number of objects in our campus dataset. Bi = bicyclist, Ped = pedestrian, Skate = skateboarders.}\label{mapcampus2}
\end{table}

% \begin{figure*}[t]
% \begin{center}
%   \includegraphics[width=.8\linewidth]{figures/pipeline}
% \end{center}
%   \caption{Illustration of our learning method. On the left hand-side, we present our learning phase. Given a large collection of raw trajectories, we cluster them into behavioral group given our social-awareness feature. Then, for each behavioral group, we learn the parameters of the corresponding forecasting model (SF stands for Social Force model). At test time, the right hand-side, we classify each trajectory into one of the group and forecast its behavior accordingly.}
% \label{fig:pipeline}
% \end{figure*}

\section{Forecasting in multi-class settings}
Our new collected dataset creates the opportunity to study methods for trajectory forecasting and evaluate them on a broader setting, \textit{i.e.} a crowded space occupied by several classes of objects. We aim to reason on the navigation style of the targets to accurately
predict their behavior; a bicyclist does not navigate the same way as a pedestrian. Even two instances from the same physical class might have different motion property given their character: some pedestrians prefer to walk fast nearby people and others prefer to stay away. We propose a unified framework that, given the observed data (short trajectories), identifies the regime (navigation style) in which it is operating to forecast the future state. We model these navigation styles as a latent variable and learn them from the data. 

%We optimize our forecasting model by inferring every target class at each time frame to  adjust every target's parameters accordingly.\\ 

\subsection{Problem formulation}

Given the observed trajectories of several targets at time $t$, we forecast their future positions over the next $N$ time frames (where $N$ is in seconds). We model the problem as a multi-class forecasting problem where the navigation style is a latent variable. 

We define the navigation style of a target as its motion properties driven by its \textit{social sensitivity} (more details in Sec. \ref{reckless}). Note that we do not use the ground truth physical class of the target since we aim to learn navigation behaviors that go beyond the physical class of an object. As a reminder, two distinct classes (\textit{e.g.} a bicyclist and skateboarder) might share the same navigation style whereas two instances from the same class (\textit{e.g.} two pedestrians) might have different styles.

%A crowd is composed of several types of moving objects (including pedestrians, bicyclists, skateboarders, or carts), but even by considering a single type of object (e.g., pedestrians), we can identify several groups of dynamics according to their moves or behaviors (passive, aggressive, or abnormal to name a few). 

% Figure \ref{fig:pipeline} summarizes our method. 
At training time, we cluster the trajectories given their navigation style, \textit{i.e.}, their \textit{social sensitivity} feature. For each cluster, we learn the parameters of our forecasting model. 
At testing time, we classify each target into one of the learned navigation style and forecast their behavior accordingly. Note the same target can have different navigation style across time.

In the rest of this section, we first present existing forecasting models and their multi-class formulation. Then, we present our forecasting framework modeling the navigation style as a hidden state.

\subsection{Forecasting model}\label{momodel}

Given a navigation style, we use the popular Social Forces model \cite{yamaguchi2011you} to forecast the target trajectories. In this section, we introduce  the basic theory behind the model and how to adapt it to multi-class settings. The model is also our inspiration for our \textit{social sensitivity} feature described in Sec. \ref{reckless}.

\paragraph{Social Forces} In this model, each target is viewed as a decision making agent who consider a multitude of personal, social and environmental factors to decide where to go next.

Each target makes a decision on its velocity $\mathbf{v}_i^{(t+\Delta t)}$. At each time step $t$, the target $i$ is defined by a state variable $ s_i^{(t)} = \left\{ \mathbf{p}_i^{(t)},\mathbf{v}_i^{(t)},u_i^{(t)},\mathbf{g}_i^{(t)},A_i^{(t)} \right\}$, where $\mathbf{p}_i^{(t)}$ is the position, $ \mathbf{v}_i^{(t)}$ the velocity, $u_i^{(t)}$ the preferred speed (according to the class and the past velocities), $\mathbf{g}_i^{(t)}$ the chosen destination (or goal) and $A_i^{(t)}$ is the set of targets in the same social group (including $i$).
Similar to \cite{yamaguchi2011you}, the energy function, $E_{\Theta}$, associated to every single target is  defined as:

\begin{equation} \label{eq1}
\begin{split}
E_{\Theta}(\mathbf{v};s_i, \mathbf{s_{-i}}) = &\lambda_0 E_{damping}(\mathbf{v};s_i) +\\
&\lambda_1 E_{speed}(\mathbf{v};s_i)  + \\
&\lambda_2 E_{direction}(\mathbf{v};s_i) +  \\
& \lambda_3 E_{attraction}(\mathbf{v};s_i,\mathbf{s}_{A_i}) + \\
 & \lambda_4 E_{group}(\mathbf{v};s_i,\mathbf{s}_{A_i})+\\
 & E_{collision}(\mathbf{v};s_i,\mathbf{s_{-i}}|\sigma_d,\sigma_w,\beta)
 \end{split}
 \end{equation}
 where $ \Theta = \left\{ \lambda_0, \lambda_1, \lambda_2, \lambda_3, \lambda_4, \sigma_d, \sigma_w, \beta \right\}$ is the model parameters,  $\mathbf{s}_{A_i}$ is the set of state variables of the target in $i$'s social group $A_i$, and  $\mathbf{s}_{-i}$ is the set of states of other targets except $i$. The parameter $\lambda_i$ are then weights to balance the importance of each of those energies ($E_.$). More details on the definition of each of the energy can be found in \cite{yamaguchi2011you}. In our work, we use the collision energy, $E_{collision}$ to define our \textit{social sensitivity} feature in Sec. \ref{reckless}. Consequently, we will describe the parameters $\left\{ \sigma_d, \sigma_w, \beta \right\}$ in Sec. \ref{reckless}.\\
 
Previous work \cite{yamaguchi2011you,pellegrini2010improving,luber2010people} use only one set of parameters for all the targets. This approximation then implies that everyone would have the same safety distance or would grant the exact same weight to every one of these energies. We can easily see that even from a simple physical constraint, a bicyclist could not have the same damping as a pedestrian, or that someone in a hurry would be more likely to bump into or navigate close to others in order to navigate faster, granting more weight to his damping energy in order to go as straight as possible to his destination.

\paragraph{Multi-class Social Forces}

We adapt the Social Forces model from single class to multiple classes. Among all the different energies to minimize in $E_{\Theta}$, some of them are specific to each object (damping, speed, direction), but some others result from the interactions with other objects, and then other classes. It is then natural to consider that the interaction between two objects of the same class (\textit{i.e.}, pedestrian-pedestrian) would not be the same as the one between two different classes (\textit{i.e.}, pedestrian-bicyclist). We will then introduce these differences by adapting the interaction energies (attraction, grouping, collision) to every single class.

At each time step $t$, we reevaluate the state variable $s_i^{(t)}$ by introducing the class $c_i^{(t)}$ associate to target $i$: $ s_i^{(t)} = \left\{ \mathbf{p}_i^{(t)},\mathbf{v}_i^{(t)},u_i^{(t)},\mathbf{g}_i^{(t)},A_i^{(t)}, c_i^{(t)} \right\}$. $\Theta $ is now re-define such as $ \Theta = \left\{ \lambda_0(c), \lambda_1(c), \lambda_2(c), \lambda_3(c), \lambda_4(c), \sigma_d(c), \sigma_w(c), \beta(c) \right\}$ where every parameter is now learned for every class considered in the map.
 
%The energies function presented in  have now their associated parameters specifically to each class considered such that:
%\begin{align}E_{\Theta}(\mathbf{v};s_i, \mathbf{s_{-i}}) = &\lambda_0(c) E_{damping}(\mathbf{v};s_i) +\\&\lambda_1(c) E_{speed}(\mathbf{v};s_i)  + \\&\lmbda_2(c) E_{direction}(\mathbf{v};s_i) +  \\
%& \lambda_3(c) E_{attraction}(\mathbf{v};s_i,\mathbf{s}_{A_i}) + \\ & \lambda_4(c) E_{group}(\mathbf{v};s_i,\mathbf{s}_{A_i})+\\
% & E_{collision}(\mathbf{v};s_i,\mathbf{s_{-i}}|\sigma_d(c),\sigma_w,\beta)
%\end{align} with \begin{itemize}
%\item $\mathbf{s}_{A_i}$ is a set of state variables of the target in $i$'s social group $A_i$.
%\item $\mathbf{s}_{-i}$ set of states of other targets except $i$.
%\end{itemize}

The optimal parameters are learned by fitting the energy function to fully observed trajectories in the labeled training data. By denoting the ground truth data with $\tilde{s}_i$,  the learning problem is defined as follows:

\begin{equation} \label{eq2}
\begin{split}
\Theta^* =\argmin_{\Theta}\sum_{i}\sum_{t}\left\vert \tilde{{\bf p}}_{i}^{(t+\Delta t)}-{\bf p}_{i}^{(t+\Delta t)}(s_{i}^{(t)},\tilde{{ s}}_{-i}^{(t)}, \Theta)\right\vert
\end{split}
\end{equation}
We can also see that most of these parameters don't need to be re-evaluated, such as the weight on the grouping or attraction energy. Our main concern would be finally to consider those different classes in a collision avoidance scene and readjust our collision parameters accordingly to the new estimated classes.

\subsection{Forecasting with a latent variable}\label{hmm}
Each target in a real-world scene belongs to a given navigation style. In reference to Sec. \ref{momodel}, a navigation style is defined by a set of unique parameters for a forecasting model, \textit{e.g.} $\Theta$ for social forces model.  We propose to learn these hidden classes from the data. Formally, we formulate the problem with a Hidden Markov Model (HMM), where observations are the short trajectories, the hidden state the navigation style, and the outputs the set of parameters, $\Theta$, for a given style.

\begin{figure}[t]
\begin{center}
   \includegraphics[width=.7\linewidth]{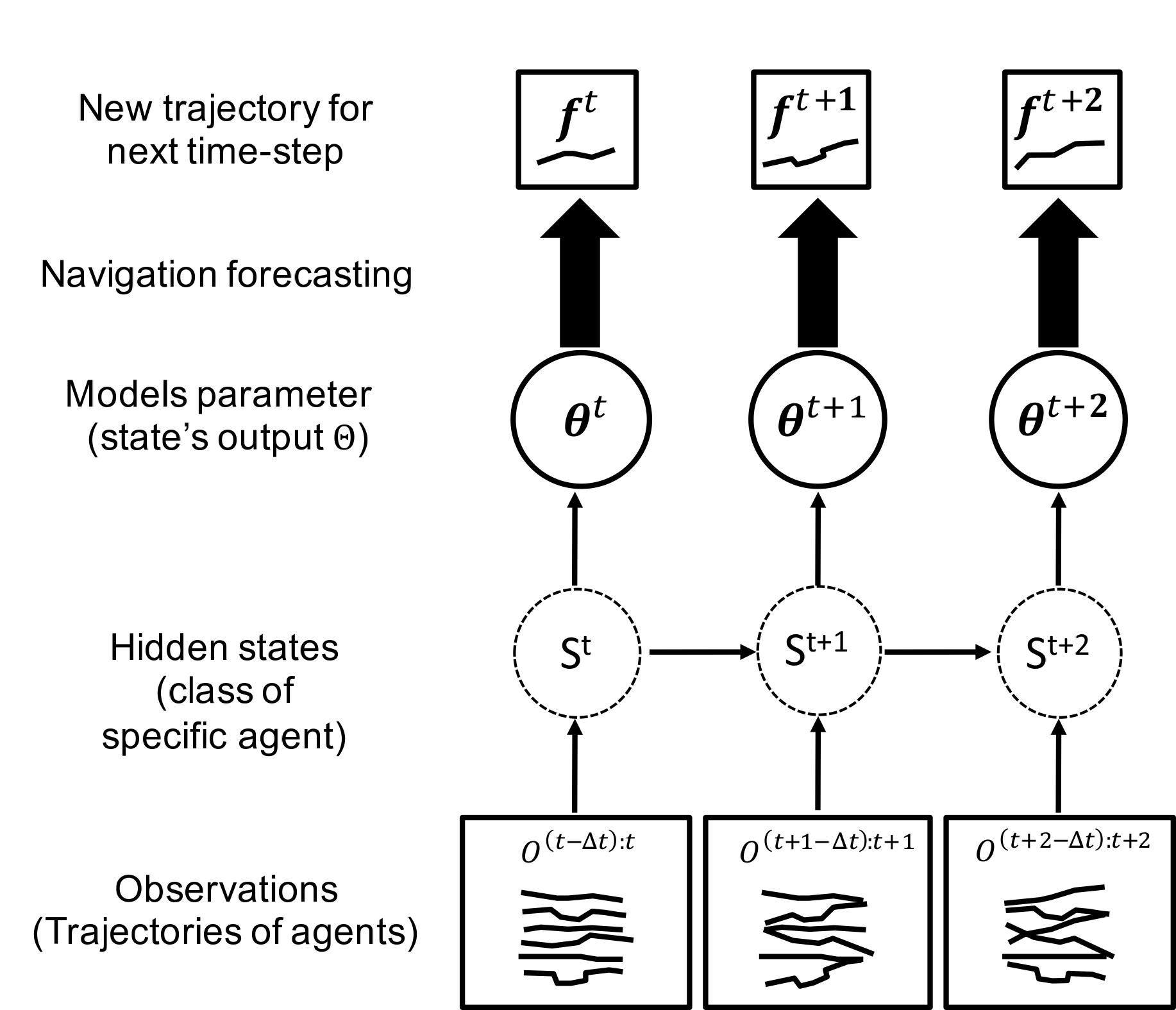}
\end{center}
   \caption{Graphical model representation of our model}
\label{hmm_fig}
\end{figure}

\paragraph{States, outputs and observations}
Similar to what described in Sec. \ref{momodel} (social forces section), in our model at each time step $t$, the state of target $i$ is defined by a state variable $ s_i^{(t)} = \left\{ \mathbf{p}_i^{(t)},\mathbf{v}_i^{(t)},u_i^{(t)},\mathbf{g}_i^{(t)},A_i^{(t)} \right\}$, where $\mathbf{p}_i^{(t)}$ is the position, $ \mathbf{v}_i^{(t)}$ the velocity, $u_i^{(t)}$ the preferred speed (according to the class and the past velocities), $\mathbf{g}_i^{(t)}$ the chosen destination (or goal) and $A_i^{(t)}$ is the set of objects in the same social group (including $i$).
The output of each state is the navigation forecasting parameters $ \Theta = \left\{ \lambda_0(c), \lambda_1(c), \lambda_2(c), \lambda_3(c), \lambda_4(c), \sigma_d(c), \sigma_w(c), \beta(c) \right\}$ where every parameter is same for every target in same class $c$.
We define target $i$, and its trajectory $\mathbf{f}_{1:T}^{(i)}=\left( \mathbf{f}^{(i)}(1),..,\mathbf{f}^{(i)}(T) \right)$ over $T$ time-steps, where each $\mathbf{f}^{(i)}(t) = (x(t),y(t)) \in \mathbb{R}^2$ is the planar location of target $i$ at time $t$.
Observation $\mathbf{O}_{i}^{t}$ is the trajectories of all the targets over the past time-steps $1$ to $t$, or $(f_i^{t-\Delta t:t},\mathbf{f_{-i}^{t-\Delta t:t}})$ where $(f_i^{t-\Delta t:t}$ is trajectory of the i-th target from time $t-\Delta t$ to t and $\mathbf{f_{-i}^{t-\Delta t:t}}$ is trajectories of all the other targets except i from time $t-\Delta t$. Figure \ref{hmm_fig} shows the graphical representation of our HMM.

\paragraph{State transitions and output probability}
Given the observations $O^{t-\Delta t:t}$ we consider that the target chooses the best move in order to minimize the social force energies, and therefore the next trajectory for each target is calculated. This is described in more details in section \ref{reckless}. 
By knowing the state of target $i$ at each time-step the output $\Theta$ for the next state is uniquely found as $\Theta(S_i^t)$.
The output probability distribution $P(\Theta^t | S_i^t=j, O^{t-\Delta t:t})$ where $S_i^t$ is the state of target $i$, is a finite mixture of Gaussian distributions associated with state $S_i^t$ and $O^{t-\Delta t:t}$.

\begin{equation} \label{eq3}
\begin{split}
P(\Theta^{t} | S_i^t=j, O^{t-\Delta t:t}) =\\ \sum_{m=1}^{M} c_{jm} G(SF_{i}(O^{t-\Delta t:t},\Theta^t),\mu_{jm},\Sigma_{jm})
\end{split}
\end{equation}

Here $c_{jm}$ is weighting coefficient for the m-th mixture in state $j$ , and $G$ is the Gaussian distribution with mean vector $\mu_{jm}$ and covariance matrix $\Sigma_{jm}$ for the m-th mixture component in state $j$. The final output of hidden state $j$ is the expected value of $\Theta$ at time $t$.
\begin{equation} \label{eq4}
\begin{split}
    E[\Theta^t | S^t=j] = E_{S^t=j}[\Theta^t] =  \\ \sum_{i=1}^{K} SF_{i}(O^{t-\Delta t:t},\Theta^t))*P(\Theta^{t} | S_i^t=j, O^{t-\Delta t:t})
\end{split}
\end{equation}
where $K$ is the number of navigation styles of targets.

\section{Learning and inferring navigation style}\label{reckless}

We claim that a single class model is not suitable for capturing the variety within the dynamic of the targets. We believe that it can be conditioned on the physical property (pedestrian, bicyclist, skateboarder, or cart) or on the character (\textit{e.g.} aggressive, or mild) of the person. We propose to learn these class of behaviors from the data by analyzing the Social Forces Energies. By considering the evolution of the different social energies introduced previously through time for every target, we can see that the real information distinguishing a target from one another is the collision energy.

We introduce a new feature, referred to as \textit{social sensitivity}, that captures the sensitivity towards others through the collision avoidance energy:

\begin{equation} \label{eq5}
\begin{split} E_{collision}(\mathbf{v};s_i,\mathbf{s_{-i}}|\sigma_d,\sigma_w,\beta) =\\  \sum_{j \neq i} w(s_i,s_j) \exp \left( - \dfrac{d^2(\mathbf{v},s_i,s_j)}{2 \sigma_d^2}\right)
\end{split}
\end{equation}
where $w(s_i,s_j)$ is a weight:

\begin{equation} \label{eq6}
\begin{split}
w(s_i,s_j) = \exp\left( - \dfrac{|\Delta \mathbf{p}_{ij} |}{2 \sigma_{\omega}} \right).\left( \dfrac{1}{2} \left(1 - \dfrac{\Delta \mathbf{p}_{ij}}{ |\Delta \mathbf{p}_{ij} |} \dfrac{\mathbf{v}_i}{|\mathbf{v_i}|} \right)\right)^{\beta}
\end{split}
\end{equation} 
and 

\begin{equation} \label{eq7}
\begin{split}
d^2(\mathbf{v},s_i,s_j)=\left\vert \Delta\mathbf{p}_{ij}-\dfrac{\Delta \mathbf{p}_{ij}(\mathbf{v}-\mathbf{v}_j)}{\vert\mathbf{v}-\mathbf{v}_j\vert^2}(\mathbf{v}-\mathbf{v}_j)\right\vert
\end{split}
\end{equation}

where $\sigma_d$ is the distance to the subject to be avoided, $\sigma_w$ the radius of influence of other objects and $\beta$ control the peakiness of the weighting function, and $d^2(\mathbf{v},s_i,s_j)$ a distance introduced in \cite{yamaguchi2011you}. We make the assumption that the radius of influence of other objects, which is related to the field of vision, is essentially the same for everyone. We also consider the parameter $\beta$ to be the same, from common sense.

Our main goal is now to evaluate for every target the parameter $\sigma_d$, \textit{i.e.}, the distance from when the target consider\textcolor{black}{s} and reacts to a potential collision. To this \textcolor{black}{extent}, we will select in our database only the scene revealing an eventual collision avoidance. This selection is realized through a combination of safety distance, field of vision, and a vector product of speed vectors. We will denote our new train and test set respectively $X_{train}^{c}$ and $X_{test}^c$.

By evaluating the distance $\sigma_d$ we interpret a very personal parameter, the average distance that a target wants to preserve from other targets on the next step. To this extent, we consider that the target chooses the best move in order to minimize its own social Force energies. Therefore, we know $v^{*(i)}(t+1)$. The idea would be then to learn how this target behaves at the approach of this collision avoidance. We then learn the parameter $\{\sigma_d(t)^i,\sigma_w(t)^i,\beta(t)^i\}$ jointly from its reaction. This parameter can be estimated at time $T$ by knowing the speed vector at time $T+1$. % Let's consider two targets $i$ and $j$. Each of this target observe and unconsciously identify the speed vector of the other one, and then predict the other target's position at the next step, considering that he won't change is trajectory.T$\sigma_d^{(i)}$ then design the average distance wanted between the target $i$ and the self-predicted position of target $j$. Of course, target $j$ having the same reaction, $\sigma_d^{(i)}$ and $\sigma_d^{(j)}$ won't be the actual distance between these two targets, but the one personally wished. 
Then, by knowing $v^{*(i)}$, at time $T$:
\small
\begin{equation}
\begin{split}
&\textcolor{black}{\{\sigma_d(T)^i,\sigma_w(T)^i,\beta(T)^i\}}\\ &= \argmin_{\textcolor{black}{\{\sigma_d,\sigma_w,\beta\}}}\left(\dfrac{1}{T}\sum_{t=1}^{T-1} E_{collision}(\textcolor{black}{\sigma_d,\sigma_w,\beta},X_{train}^{c})\right)\\ 
&=\argmin_{\textcolor{black}{\{\sigma_d,\sigma_w,\beta\}}}\left(\dfrac{1}{T}\sum_{t=1}^{T-1}\sum_{j \neq i} w(t,\textcolor{black}{\sigma_w,\beta}) \exp \left( - \dfrac{d^2(\mathbf{v^{*(i)}}(T+1))}{2 \sigma_d^2}\right)\right)
\end{split}
\end{equation}
with 
$$ w(t,\textcolor{black}{\sigma_w,\beta}) = w(s_i(t,\textcolor{black}{\sigma_w}),s_j(t,\textcolor{black}{\beta})) $$and$$  d^2(\mathbf{v^{*(i)}}(t+1)) = d^2(\mathbf{v}^{(i)}(t+1),s_i(t),s_j(t))$$

\normalsize

This minimization is operated with an interior-point method and setting the following constraint on $\sigma_d$: \hspace{.5cm} $\sigma_d >0.1$ (specifying that every target can't have a ``vital space'' less than $10$cm). The fact that $w(t,\sigma_w,\beta)$ is in itself sufficient to avoid $\sigma_d$ being null.

The distance $\sigma_d$ is then identified for each target in our training set. In order to model the change in behaviors with respect to crowd density, we include the traffic density in the modeling of a target. We define the \textit{social-sensitivity} of a target $(i)$ at time $t$ with the parameter $\sigma_{sa}$  such as: 
\begin{equation}
\sigma_{sa}^{(i)}(t)=\exp\left(-\sigma_d^{(i)}/ds^2(t)\right)
\end{equation}
with $ds(t)$ the density at time $t$. \\
We clustered our training set with a k-means algorithm and evaluate our clusters through our collision-test set. The optimal number of clusters is established with a Calinski-Harabasz criterion clustering evaluation \cite{muhr2009automatic}. Once those clusters are identified, we label every target in our training set according to it and train a classifier (a SVM in our experiment) using the (speed, position, damping, and \textit{social-sensitivity}).

In order to consider only the relevant values of this new feature, in our training set we will select only the frames containing a potential collision scene. By observing the target reaction (speed and trajectory), we will calculate $\sigma_{sa}$ as the value minimizing the collision energy. %We propose to infer the label of forecasting models given the observed velocity and interaction between targets. To accomplish this, w
We train an SVM classifier for each collision scene and pedestrian for whom the $\sigma_{sa}$ is known, to then predict the target's label and select the set of parameters to use. This way, the different behavior of each target would affect only the collision avoidance problem, and could also change through time and situation.\\

% \begin{figure*}[ht!]
% 		\begin{center}
% 			\includegraphics[width=1\linewidth]{figures/qual_res.pdf}
% 		\end{center}
% 		\caption{Illustration of qualitative results on Campus dataset. In green the ground truth trajectory of the target, in orange the Kalman Linear prediction, in blue the predicted trajectory given social force model, and in red the predicted trajectory given our SF-sa model.}
% 		\label{fig:stanford}
% \end{figure*}

% \vspace{-3mm}

\section{Experiments}

\subsection{Datasets and metrics}
We evaluate our multi-class forecasting framework on our new collected dataset as well as previous existing pedestrian-only ones \cite{pellegrini2009you,lerner2007crowds}. Our dataset as two order of magnitude more targets than the combined pedestrian-only datasets. We evaluate the performance of forecasting methods with the following measures: average prediction error over (i) the full estimated trajectory, (ii) the final estimated point, and (iii) the average displacement during collision avoidance's. Similar to \cite{pellegrini2009you,lerner2007crowds}, we observe trajectories for 2.4 seconds and predict for 4.8 seconds. We sub-sample a trajectory every 0.4 second. We also focus our evaluation when non-linear behaviors occur in the trajectories to not be affected by statistically long linear behaviors.

\subsection{Navigation style assignments}

\begin{figure}[!t]
\begin{center}
   \includegraphics[width=0.75\linewidth]{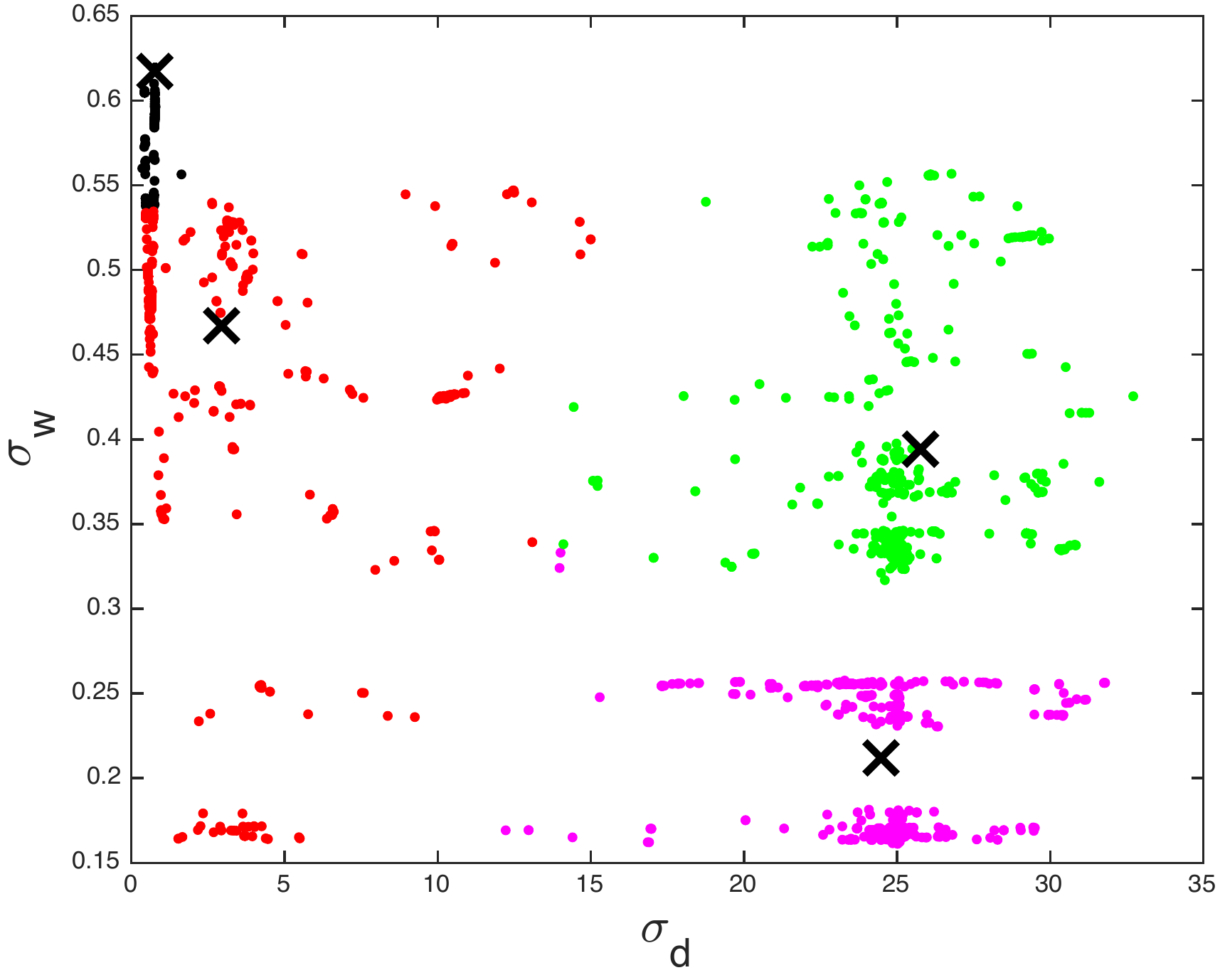}
\end{center}
   \caption{Illustration of the \textit{social sensitivity} space. Each point is a target. The the x-axis is the distance at which a target starts deviating from its linear trajectory in order to avoid an upcoming collision, and y-axis is the average distance a target keeps with its surrounding targets (average personal distance to other targets).  Each color code represents a cluster (a navigation style). Even if our approach can handle an arbitrary number of classes, we only use 4 clusters for illustration purposes. In the remaining figures \ref{fig:sfig21} to \ref{fig:sfig33}, we use the same color convention for each navigation style. In this plot, the green cluster represents targets with a mild behavior, willing to avoid other targets as much as possible and considering them from afar, whereas the red cluster describes targets with a more aggressive behavior and with a very small safety distance, considering others at the last moment.}
\label{fig:sfig2c}
\end{figure}

\begin{figure*}[t]
\begin{center}
   \includegraphics[width=0.68\linewidth]{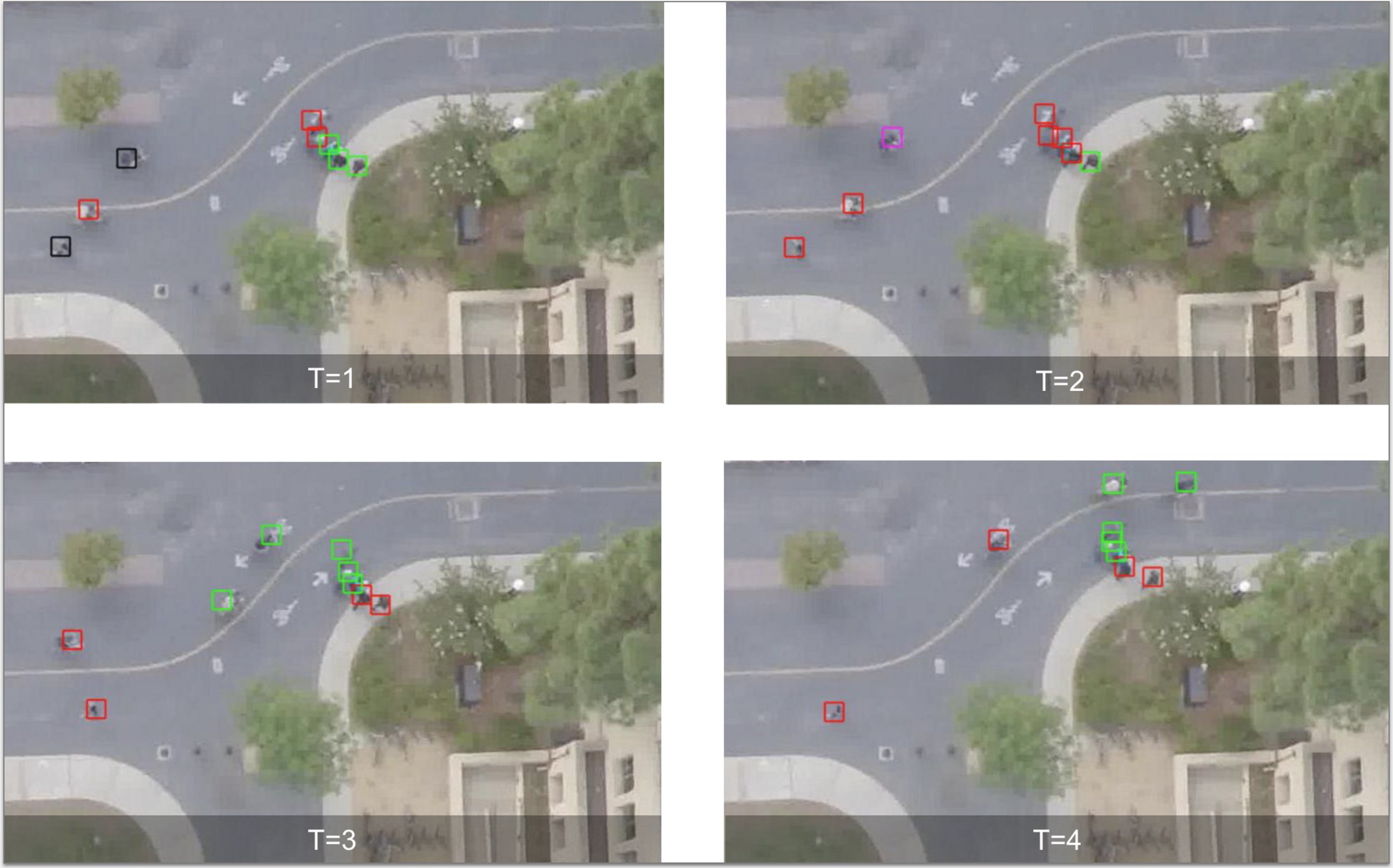}
\end{center}
   \caption{Illustration of the class assignment for each target. The same color represents the same navigation style (cluster) described in Figure \ref{fig:sfig2c}. Note that for a given target its class changes across time regardless of its physical class. When the target is surrounded by other targets, its class changes with respect to its \textit{social sensitivity}. In this scene, first we can observe a cyclist (shown as label 1 in the images) belonging to a black cluster, \textit{i.e.}, being aggressive in his moves, then belonging to some milder clusters (purple and green). We also can see the evolution of a group of pedestrians  (shown as labels 2,3) in the images), initially ``mild'' (green at $T=1$), who become red at time $T=3$ at which they decide to overtake another group and accelerate.}
\label{fig:sfig21}
\end{figure*}

\begin{figure*}[t]
\begin{center}
   \includegraphics[width=0.68\linewidth]{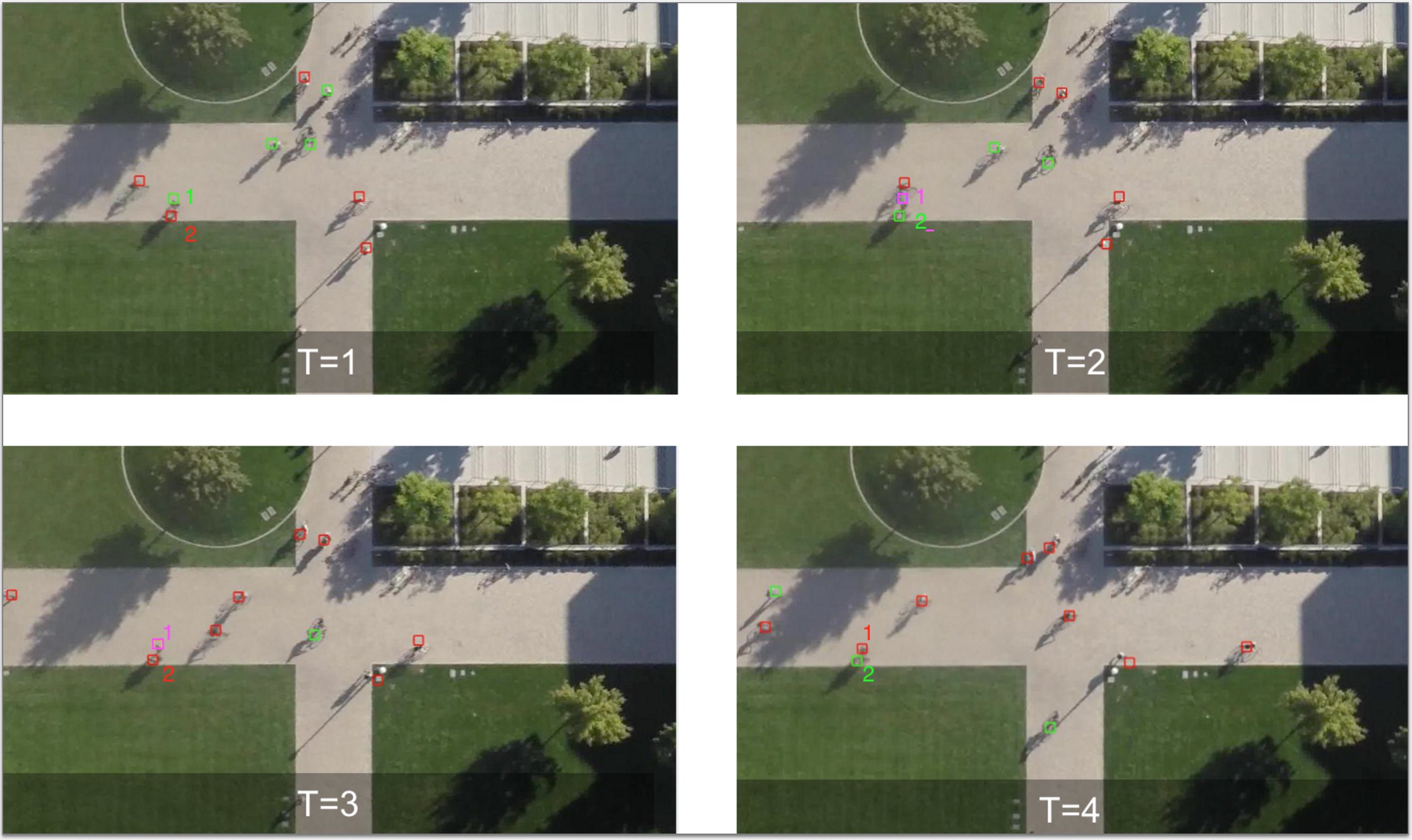}
\end{center}
   \caption{Illustration of the class assignment for each target. Same color represents the same navigation style (cluster) described in Figure \ref{fig:sfig2c}. Note that for a given target its navigation style changes across time regardless of its physical class.  When the target is surrounded by other targets, its class changes with respect to its \textit{social sensitivity}. In this plot, we can see a target  (shown as label 1 in the images) being first green ($T=1$) then purple during a collision avoidance, describing a "last minute" but ``mild'' reaction, leading it to slow down, to turn red in order to catch up on its partner  (shown as label 2 in the images).}
\label{fig:sfig22}
\end{figure*}

In Sec. \ref{reckless}, we introduce our \textit{social sensitivity} feature to cluster and classify trajectories into different navigation style. Figure \ref{fig:sfig2c} plots each target into the \textit{social sensitivity} space where the x-axis is the distance at which a target starts deviating from its linear trajectory in order to avoid an upcoming collision, and y-axis is the average distance a target keeps with its surrounding targets (average personal distance to other targets). Each cluster corresponds to a navigation style. A navigation style describes the sensitivity of a target to its surrounding. It is different than its physical class such as a pedestrian or bicyclist. In the absence of interactions, a target takes either a default navigation style (when entering a scene) or the last inferred class during the previous interaction. The default navigation style is the most popular one (in red in Figure \ref{fig:sfig2c}). In figures \ref{fig:sfig21} and \ref{fig:sfig22}, we present the navigation style of each target predicted using the \textit{social sensitivity} feature. When the target is surrounded by other targets, its class changes with respect to its \textit{social sensitivity}.\\

\small
\begin{table*}[!t]
\footnotesize{
\begin{center}
\begin{tabular}{|c|c|c|c|c|c|}
\hline Methods &Lin& LTA &SF \cite{yamaguchi2011you} &  IGP \cite{trautman2010unfreezing} &  Our SF-sa   \\ 
\hline \textsc{eth} & 0.80 $|$ 0.95 $|$ 1.31 & 0.54 $|$ 0.70 $|$ 0.77 & 0.41 $|$ 0.49 $|$ 0.59  &\textbf{0.20} $|$ \textbf{0.39} $|$ \textbf{0.43} & 0.41 $|$ 0.46 $|$ 0.59 \\ 
\hline \textsc{Hotel}& 0.39 $|$ 0.55 $|$ 0.63 &0.38 $|$ 0.49 $|$ 0.64 & 0.25 $|$ 0.38 $|$ 0.37 &\textbf{0.24} $|$ 0.34 $|$ 0.37&\textbf{0.24} $|$ \textbf{0.32} $|$ \textbf{0.37}\\
\hline \textsc{Zara 1}& 0.47 $|$ 0.56 $|$ 0.89 & 0.37 $|$ 0.39 $|$ 0.66 & 0.40 $|$ \textbf{0.41} $|$ 0.60 &  0.39  $|$ 0.54  $|$ \textbf{0.39} & \textbf{0.35} $|$  \textbf{0.41}  $|$ 0.60 \\
\hline \textsc{Zara 2} & 0.45 $|$ 0.44 $|$ 0.91 & 0.40 $|$ 0.41 $|$ 0.72 & 0.40 $|$ 0.40 $|$ 0.68 & 0.41  $|$ 0.43 $|$ 0.42 & \textbf{0.39} $|$ \textbf{0.39} $|$ 0.67\\ 
\hline \textsc{UCY} & 0.57 $|$ 0.62 $|$ 1.14 & 0.51 $|$ 0.57 $|$ 0.95 & 0.48 $|$ 0.54 $|$ 0.78 & 0.61 $|$ 0.62 $|$ 1.82 & \textbf{0.45} $|$ \textbf{0.51} $|$ \textbf{0.76}\\ 
%\hline \textsc{Campus} & 0.37 &\textcolor{black}{0.35}&\textbf{ 0.34} & \textbf{0.34} & 0.672 \\%0.59 
\hline 
\hline \textsc{Average} & 0.54 $|$ 0.62 $|$ 0.97 & 0.44 $|$ 0.51 $|$ 0.75 & 0.39 $|$ 0.44 $|$ \textbf{0.60} & \textbf{0.37} $|$ 0.46 $|$ 0.69 & \textbf{0.37} $|$ \textbf{0.42} $|$ \textbf{0.60}\\ 
%\hline \textsc{Campus} & 0.37 &\textcolor{black}{0.35}&\textbf{ 0.34} & \textbf{0.34} & 0.672 \\%0.59 
\hline 
\end{tabular}

\caption{Pedestrian Only dataset - Our 3 main evaluation methods, ordered as: Mean Average Displacement on all trajectories $|$ Mean Average Displacement on collisions avoidance $|$ Average displacement of the predicted final position (after 4.8 seconds).} 
\label{pedanalysis}
\end{center}
}
\end{table*}

\small
\begin{table*}[!t]
\footnotesize{
\begin{center}
%\footnotesize{
\begin{tabular}{|c|c|c|c|c|c|}
\hline Methods &Lin&SF&  IGP \cite{trautman2010unfreezing} &  SF-Physical & Our SF-sa   \\ 
\hline \textsc{Isengard} & 1.69 $|$  1.00 $|$ 2.84 & 1.60 $|$ 0.99 $|$ 2.32 & 1.57 $|$ 1.14 $|$ 2.64 & 1.56 $|$ 0.86 $|$ 1.83 & \textbf{1.53} $|$ \textbf{0.84} $|$ \textbf{1.81} \\ 
%\hline \textsc{Little}  & $|$  $|$ &   $|$  $|$ & $|$  $|$ & $|$  $|$  \\ 
\hline \textsc{Hobbiton} &1.17 $|$ 1.01 $|$ 1.81 & \textbf{1.11} $|$ 0.82 $|$ 1.70 & \textbf{1.11} $|$  \textbf{0.81} $|$ 2.25 & 1.12 $|$ \textbf{0.81} $|$ \textbf{1.70} & 1.12 $|$ 0.83 $|$ \textbf{1.70}\\ 
\hline \textsc{Edoras}  & 0,91 $|$ 0.83 $|$ 1.03 & 0.80 $|$  \textbf{0.81} $|$ 0.89 & 1.33 $|$ 0.85 $|$ 2.61 & 0.79 $|$  \textbf{0.81} $|$ \textbf{0.89} & \textbf{0.78} $|$  0.82 $|$ \textbf{0.89} \\ 
\hline \textsc{Mordor}  & 1.72 $|$ 1.10 $|$ 3.80 & 1.38 $|$ 0.89 $|$ 2.30 & \textbf{0.95} $|$ 0.69 $|$ \textbf{1.78} & 1.37 $|$ 0.65  $|$  2.30 & 1.37 $|$ \textbf{0.60}  $|$  2.30\\ 
%\hline \textsc{Main Quad}  &1.17 $|$ 1.01 $|$ 1.8& 1.11$|$ 0.82 $|$1.7&$|$$|$&$|$ $|$\\
\hline \textsc{Fangorn}  & 1.02 $|$ 0.75 $|$ 2.00 & 0.94 $|$ 0.41  $|$ 1.66 & 0.96 $|$ 0.69 $|$ 1.67 & 0.90 $|$ 0.40 $|$ \textbf{1.51}  & \textbf{0.89} $|$ \textbf{0.36} $|$ \textbf{1.51}  \\ 
\hline \textsc{The Valley} & 1.38 $|$ 0.86 $|$ 2.45 & 1.29 $|$ 0.87 $|$ 2.02 & 1.20 $|$ 0.75 $|$ 2.46 & 1.01 $|$ \textbf{0.65} $|$ \textbf{1.65} & \textbf{0.99} $|$ 0.66 $|$ \textbf{1.65} \\ 
\hline
\hline \textsc{Average} & 1.32 $|$ 0.93 $|$ 2.32 & 1.29 $|$ 0.79 $|$ 1.82 & 1.19 $|$ 0.82 $|$ 2.24 & 1.14 $|$ 0.70 $|$ 1.65 & \textbf{1.11} $|$ \textbf{0.69} $|$ \textbf{1.64} \\ 
%\hline \textsc{Bytes} & $|$  $|$ &  $|$  $|$ & $|$  $|$ & $|$  $|$  \\ 
%\hline \textsc{Campus} & \textbf{\textcolor{black}{20458}} & \textbf{\textcolor{black}{752}} & \textbf{\textcolor{black}{28250}}&\textcolor{black}{\textbf{7}}\\ 
\hline 
% }
\end{tabular}
\caption{Campus Dataset - Our 3 main evaluation methods, ordered as: Mean Average Displacement on all trajectories $|$ Mean Average Displacement on collisions avoidance $|$ Average displacement of the predicted final position (after 4.8 seconds).}
\label{campusanalysis}
\end{center}
}
\end{table*}
\normalsize

\subsection{Forecasting accuracy}

We evaluate our proposed multi-class forecasting framework against the following baselines: (i) single class forecasting methods such as SF \cite{yamaguchi2011you} and IGP \cite{trautman2010unfreezing}, (ii) physical class based forecasting (SF-pc), \textit{i.e.}, using the ground truth physical class, and (iii) our proposed method inferring navigation style of the targets referred to as SF-sa. 

We present our quantitative results in Tables \ref{pedanalysis} and \ref{campusanalysis}: 
\textbf{On pedestrian-only dataset} (Table \ref{pedanalysis}), our SF-sa performs the same as the single class Social Forces model in \textsc{ETH} dataset, and outperforms other methods in UCY datasets. This result can be justified by the fact that the UCY dataset is considerably more crowded, with more collisions, and therefore presenting different types of behaviors. Non-linear behaviors such as people stopping and talking to each other, walking faster, or turning around each others are more common in UCY than in ETH. Our forecasting model is able to infer these navigation patterns hence better predict the trajectories of pedestrians.  

We also report the performance of the IGP model on these pedestrian-only datasets for completeness. Its accuracy is not better than Social Forces in crowded settings although it uses the destination and time of arrival as additional inputs. 

\textbf{On our multi-class dataset} (Table \ref{campusanalysis}), we can see that our approach is more accurate on every scenes containing a large amount of different classes. Our highest gain in performance is visible on the last three groups of scenes, rich in classes and collisions (see Table \ref{mapcampus}). In \textsc{Hobbiton} and \textsc{Edoras} scenes, our algorithm, trained on a multiclass dataset, matches the single class Social Forces. This happens because the latent variable of our HMM is updated to one of the classes at each collision. In a scene with less number of classes, this could become a drawback, but yet our algorithm can perform with the same accuracy.

Table \ref{campusanalysis} also compares the performance of using our method against a multi-class approach using the physical classes (limiting the SF-sa to only consider the physical classes and not adding any more classes in unsupervised way), \textit{i.e.}, one model parameter per physical class (referred to as SF-Physical). Note that both multi-class strategies perform almost the same although our method has less information. As a reminder, our method does not  require any prior on the target such as its physical class. 

\begin{figure*}[!ht]
\begin{center}
   \includegraphics[width=.55\linewidth]{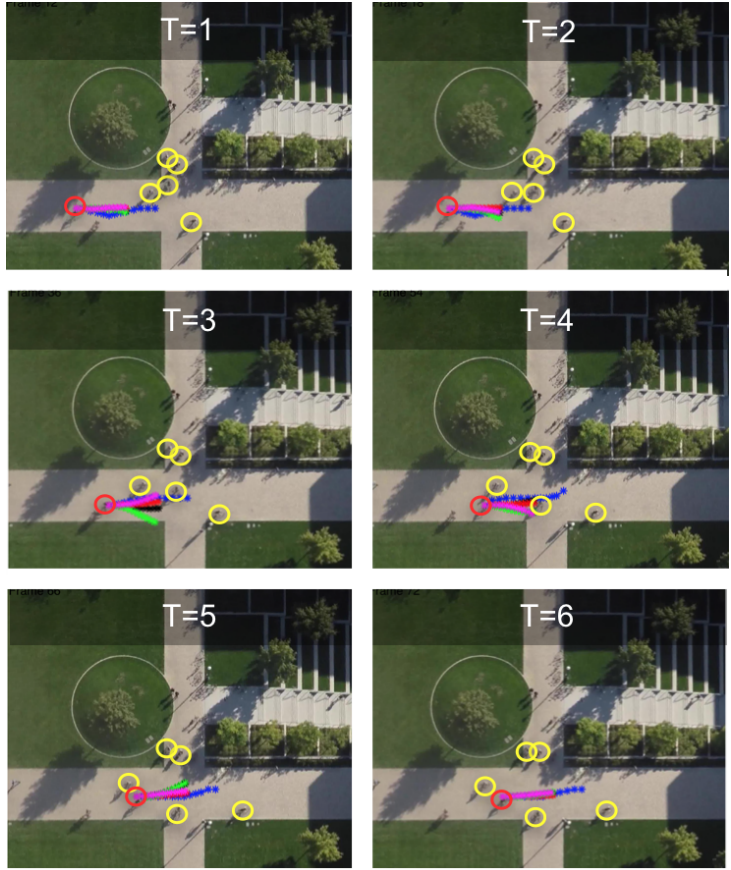}
\end{center}
   \caption{We show the predicted trajectory of a given target (red circle) in which four different navigation styles are used to perform the prediction. The corresponding predicted trajectories are overlaid on one other and shown with different color codes (the same as those used for depicting the clusters in figure \ref{fig:sfig2c}, \ref{fig:sfig21}, \ref{fig:sfig22}). The ground truth is represented in blue. Predicted trajectories are shown for 6 subsequent frames indicated by $T=1,...,6$ respectively. Interestingly, when the target is far away from other targets (no interactions are taking place) the predicted trajectories are very similar to each other (they almost overlap and show a linear trajectory).  However, when the red target gets closers to other targets (e.g. the ones indicated in yellow), the predicted trajectories start showing different behaviors depending on the navigation style: a conservative navigation style enables trajectories' prediction that keep large distances to the yellow targets in order to avoid them (green, purple trajectory) whereas an aggressive navigation style enables trajectories' prediction that are not too distant from the yellow targets (red trajectory). Notice that our approach is capable to automatically associate the target to one of the 4 clusters based on the characteristics in the “\textit{social sensitivity} space” that have been observed until present. In this example, our approach selects the red trajectory which is indeed the closest to the ground truth's predicted trajectory (in blue).}
\label{fig:sfig31}
\end{figure*}

\begin{figure*}[!ht]
\begin{center}
   \includegraphics[width=0.5\linewidth]{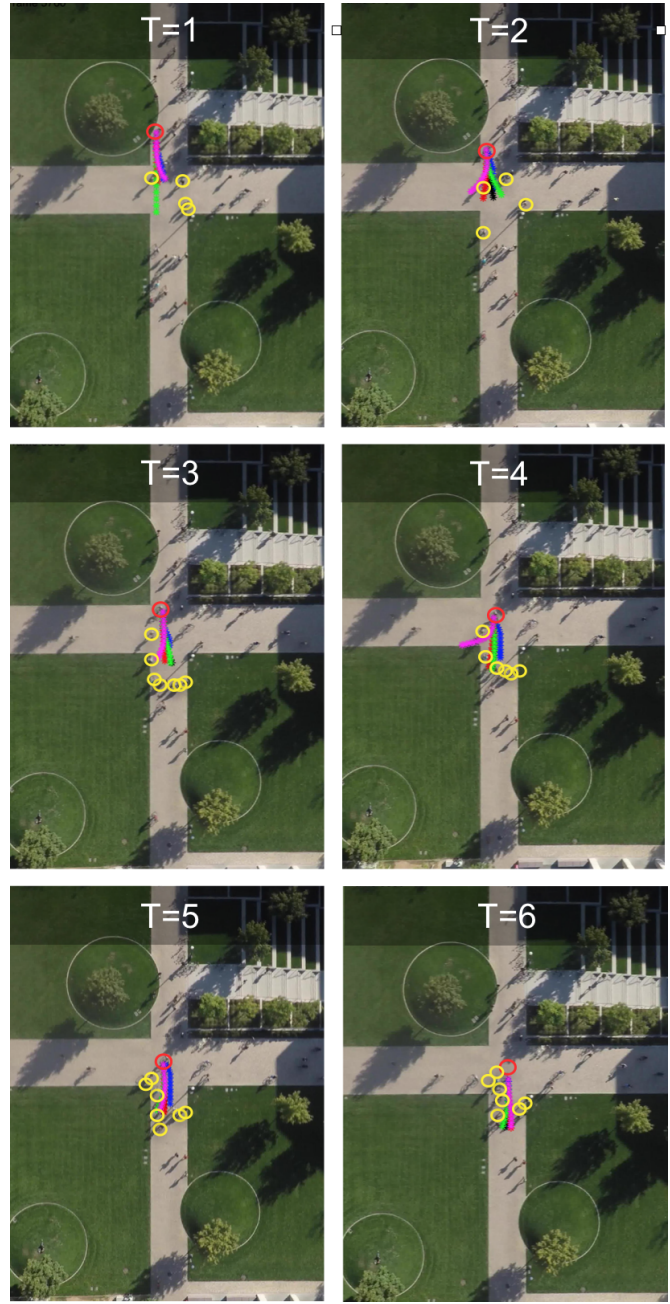}
\end{center}
   \caption{We show the predicted trajectory of a given target (red circle) in which four different navigation styles are used to perform the prediction. The corresponding predicted trajectories are overlaid on one other and shown with different color codes (the same as those used for depicting the clusters in figure \ref{fig:sfig2c}, \ref{fig:sfig21}, \ref{fig:sfig22}). The ground truth is represented in blue. Predicted trajectories are shown for 6 subsequent frames indicated by $T=1,...,6$ respectively. Interestingly, when the target is far away from other targets (no interactions are taking place) the predicted trajectories are very similar to each other (they almost overlap and show a linear trajectory).  However, when the red target gets closers to other targets (e.g. the ones indicated in yellow), the predicted trajectories start showing different behaviors depending on the navigation style: a conservative navigation style enables trajectories' prediction that keep large distances to the yellow targets in order to avoid them (green, purple trajectory) whereas an aggressive navigation style enables trajectories' prediction that are not too distant from the yellow targets (red trajectory). Notice that our approach is capable to automatically associate the target to one of the 4 clusters based on the characteristics in the “\textit{social sensitivity} space” that have been observed until present. In this example, our approach selects the red trajectory which is indeed the closest to the ground truth's predicted trajectory (in blue).}
\label{fig:sfig32}
\end{figure*}

\begin{figure*}[!ht]
\begin{center}
   \subfigure[Agressive behavior]{\includegraphics[width=0.45\linewidth]{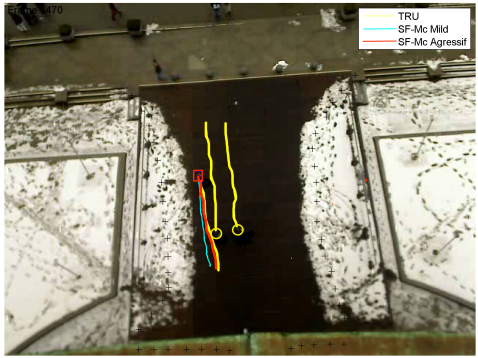}}
   \subfigure[Mild behavior]{\includegraphics[width=0.45\linewidth]{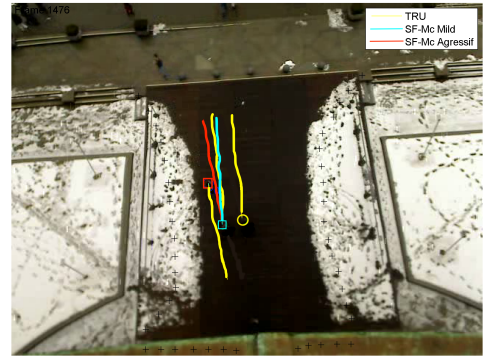}}
\end{center}
   \caption{We also evaluated our method on the pedestrian-only dataset \cite{pellegrini2009you}. We show the predicted trajectory of two pedestrians (red square and cyan square):  (a) an aggressive behavior (in the left image), and a mild one (right image). For each pedestrian, we show the predicted trajectory in which 2 different navigation styles are used to perform the prediction. The corresponding predicted trajectories are overlaid on one other and shown with two color codes: in cyan a mild behavior (preferring large distance to other pedestrians), and in red an aggressive behavior (moving forward in the close proximity of other pedestrians). The ground truth is represented in yellow. On the left image (a),  our approach selects the red trajectory which is indeed the closest to the ground truth's predicted trajectory (in yellow). On the right image (b), our approach selects the cyan trajectory which is again the closest to the ground truth's predicted trajectory. It demonstrates the relevance of our method even if a single physical class of objects interact with each other.}
\label{fig:sfig33}
\end{figure*}

We further study the impact of the number of clusters used by our method on the forecasting accuracy in Table \ref{table:num_cluster}. Once a target is associated to one of the navigation styles, the corresponding parameter $\theta$ from Equation 7 is used to predict the trajectory of the target. We can visualize the impact of the navigation style on the prediction. In figures \ref{fig:sfig31} to \ref{fig:sfig33}, we show the predicted trajectories when several navigation styles are used to perform the prediction. It is interesting to notice that when a target is far away from other targets (no interactions are taking place), all navigation styles exhibit similar linear trajectories. However, in the presence of other targets, each navigation style behaves differently. As seen in figure \ref{fig:sfig31} at time $T = 3$, there is a marked change in the trajectory of the ``green" navigation style compared to the others. This depicts a more conservative behaviour with strong repulsion to neighboring targets. This shows the need to assign targets into specific classes. All experiments results in table \ref{campusanalysis} are given considering 7 clusters. 

% In Fig. \ref{fig:cluster}, we illustrate the data points used to find these clusters. Every point in Figure \ref{fig:cluster} is defined by the set $\{\sigma_d, \sigma_w\}$, describing the specific parameters of the collision energy. The apparition of such clusters proves that people's behavior during a collision could be similar to other target's, and then clustered. For instance, a target identified as belonging to the yellow cluster would be very careful, and will try to avoid as much as possible other targets. In contrast, a target belonging to the red cluster will  walk nearby other people without respecting their private space. 

\begin{table}[!ht]
\begin{center}
\footnotesize{
\begin{tabular}{|c|c|c|c|c|c|c|}
\hline  & 1 \cite{yamaguchi2011you} & 2 & 4 & 7 & 12 & 18\\ 
\hline Mean error & 1.14&  1.16 & 1.15& \textbf{1.11} & 1.12 & 1.20 \\ 
\hline Collision error&0.72 & \textbf{0.68}& 0.69 & 0.69 & 0.73 & 0.75\\ 
\hline Final position error & 1.84 & 1.74 & 1.70 & \textbf{1.64} & 1.69 & 1.80 \\ 
%\hline \textsc{Campus} & \textbf{\textcolor{black}{20458}} & \textbf{\textcolor{black}{752}} & \textbf{\textcolor{black}{28250}}&\textcolor{black}{\textbf{7}}\\ 
\hline 
\end{tabular}
}
\caption{Forecasting error with respect to the number of clusters in our new campus dataset.}
\label{table:num_cluster}
\end{center}
\end{table}

\begin{figure*}[!ht]
\begin{center}
   \includegraphics[width=\linewidth]{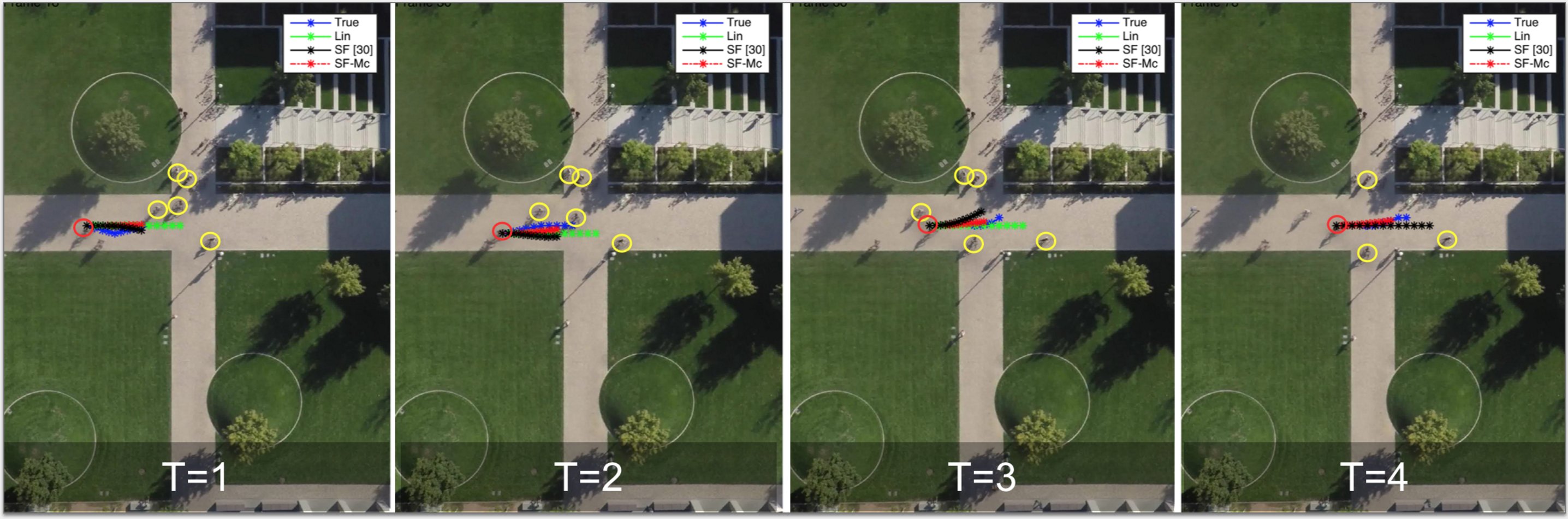}
\end{center}
   \caption{Illustration of the predicted trajectories by our SF-mc method (in red) across time. Predicted trajectories are shown for 4 subsequent frames indicated by $T=1,...,4$ respectively. We compare them with previous works (similar to Figure 4 from our paper). The ground truth is represented in blue. Our proposed multi-class framework outperforms previous methods in crowded scenes. However, in the absence of interactions, all methods perform the same. We show the same sequence as in Figure \ref{fig:sfig31}.}
\label{fig:sfig41}
\end{figure*}

\begin{figure*}[!ht]
\begin{center}
   \includegraphics[width=\linewidth]{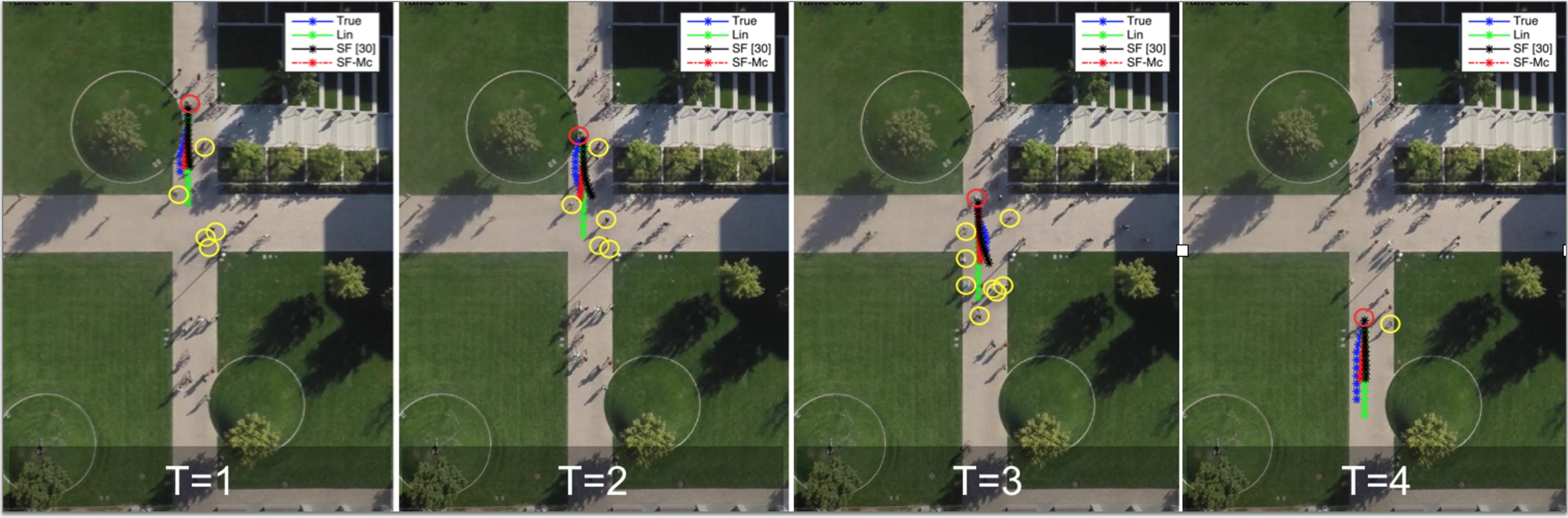}
\end{center}
   \caption{Illustration of the predicted trajectories by our SF-mc method (in red) across time. Predicted trajectories are shown for 4 subsequent frames indicated by $T=1,...,4$ respectively. We compare them with previous works (similar to Figure 4 from our paper). The ground truth is represented in blue. Our proposed multi-class framework outperforms previous methods in crowded scenes. However, in the absence of interactions, all methods perform the same. We show the same sequence as in Figure \ref{fig:sfig32}.}
\label{fig:sfig42}
\end{figure*}

Finally, in figures \ref{fig:sfig41} and \ref{fig:sfig42}, we show more examples of our predicted trajectories and compare them with previous works. Our proposed multi-class framework outperforms previous methods in crowded scenes. However, in the absence of interactions, all methods perform the same. Future work will focus on improving the forecasting performance in crowded contexts where non-linear behaviors occur.

% \begin{figure}[t]
% \begin{center}
%   \includegraphics[width=0.9\linewidth]{figures/cluster4.eps}
% \end{center}
%   \caption{Illustration of the ``social sensitivity" space (similar to Figure 5 in our paper). Each point is a target. The the x-axis is the distance at which a target starts deviating from its linear trajectory in order to avoid an upcoming collision, and y-axis is the average distance a target keeps with its surrounding targets (average personal distance to other targets).  Each color code represents a cluster (a navigation style). Even if our approach can handle an arbitrary number of classes, we only use 4 clusters for illustration purposes. In the remaining figures \ref{fig:sfig21} to \ref{fig:sfig33}, we use the same color convention for each navigation style. In this plot, the green cluster represents targets with a mild behavior, willing to avoid other targets as much as possible and considering them from afar, whereas the red cluster describes targets with a more aggressive behavior and with a very small safety distance, considering others at the last moment.}
% \label{fig:sfig2c}
% \end{figure}

\section{Conclusions}

We have presented our efforts to study human navigation at a new scale. We have contributed the first large-scale dataset that has top view videos of multiple classes of objects interacting in complex and crowded university campus. We have presented our work on predicting the trajectories of several classes of objects without explicitly solving the object classification task. Future work will study other forecasting methods such as Long Short-Term Memory (LSTM) to jointly solve the prediction task. Finally, by sharing our dataset, we hope that researchers will push the limits of existing methods in modeling human interactions, learning scene specific human motion, inferring functional maps of a scene, or detecting and tracking tiny targets from UAV data.

\raggedright
%\newpage

\cleardoublepage
\newpage
  
\newpage

%\ifodd\value{column}\hbox{}\newpage\fi
%\cleardoublepage

\footnotesize

\bibliographystyle{ieee}
\bibliography{stanfordbibli1}

\end{document}